\documentclass{article}

    \usepackage[nonatbib, preprint]{neurips_2019}

\usepackage[utf8]{inputenc} 
\usepackage[T1]{fontenc}    
\usepackage[bookmarks=false]{hyperref}
\usepackage{url}            
\usepackage{booktabs}       
\usepackage{amsfonts}       
\usepackage{nicefrac}       
\usepackage{microtype}      
\usepackage{amsmath}

\usepackage{float}
\usepackage{graphicx}
\usepackage{subfig}

\usepackage{cite}

\usepackage{fancyhdr}
\title{Mix and Mask Actor-Critic Methods}

\author{
  \textbf{Dom Huh} \\ 
  $^{1}$Department of Computer Science, \\
  University of California, Davis, CA, USA \\
  \texttt{\{dhuh\}@ucdavis.edu}
}

\begin{document}

\maketitle


\begin{abstract}
Shared feature spaces for actor-critic methods aims to capture generalized latent representations to be used by the policy and value function with the hopes for a more stable and sample-efficient optimization. However, such a paradigm present a number of challenges in practice, as parameters generating a shared representation must learn off two distinct objectives, resulting in competing updates and learning perturbations. In this paper, we present a novel feature-sharing framework to address these difficulties by introducing the mix and mask mechanisms and the distributional scalarization technique. These mechanisms behaves dynamically to couple and decouple connected latent features variably between the policy and value function, while the distributional scalarization standardizes the two objectives using a probabilistic standpoint. From our experimental results, we demonstrate significant performance improvements compared to alternative methods using separate networks and networks with a shared backbone. The code for this paper can be found in the following link: \url{https://github.com/domhuh/mixmaskac}.
\end{abstract}

\section{Introduction} 
Deep reinforcement learning (RL) has risen to accomplish impressive feats in diverse applications: from playing Atari games \cite{mnih2013playing} to solving classic control tasks \cite{DBLP:journals/corr/DuanCHSA16}. The success of this field can be attributed to breakthroughs in novel optimization methods \cite{svg, a3c, dpg, td3, TRPO, PPO, SAC} and innovative mechanisms  \cite{mnih2013playing,schaul2016prioritized,andrychowicz2018hindsight,acer} to boost stability and sample-efficiency. In particular, many of these algorithms have relied on the RL framework called the actor-critic method. As demonstrated in past literature \cite{actorcritic}, this class of algorithms introduce a bootstrapping value function working in conjunction with policy gradient optimization as to reduce variance and accelerate learning. 

In this paper, we suppose and argue the hypothesis that sharing latent features between a policy and value function will capture more generalized and rich representations and that latent features encouraged by one function may be highly useful for the other and vice-versa, enabling for more stable and sample-efficient optimization. In turn, we assert that the representations learned by both the policy and value function are complementary in some respect, thereby should be related in some manner. However, we must address two central problems that arises in this proposition: the definability of function interdependence and the challenge of multi-objective optimization.

Let us consider the dynamics of the relationship between the policy and value function. We focus on two important components that ensures a stable and definable interdependence: the initialization and conditioning of the two functions in question. Often for actor-critic algorithms, we treat the initialization and conditioning process for the policy and value function independently, and update, or condition, these functions disproportionately. Consequently, their interdependence therefore begins ambiguously and remains non-stationary throughout optimization. In fact, this connection can be quickly redefined as the application at hand changes. Hence, having shared features between these two functions can detriment learning if proper precautions are not in place since their relationship is difficult to define and can be easily distorted. Secondly, as the actor-critic methods use two distinct objectives for the policy and value function, the presence of a shared latent space forms a multi-objective optimization problem. In this setting, we are confronted with two issues. First, it is possible for one objective to dominate over the other, resulting in performance disparity between objectives. Second, there may be interference between the two objectives during optimization, causing competing changes in the parameters. With dominance and perturbations, training without any safeguard can be highly unstable and sample-inefficient.

\begin{figure}
    \centering
    \includegraphics[width=\textwidth]{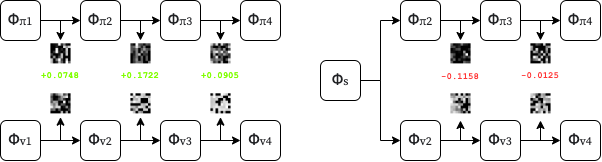}
    \caption{Change in Similarity Measure: On the left, the figure shows the separated network paradigm. On the right, the figure shows the shared backbone paradigm. For both paradigms, a visualization of the hidden representations and the expected difference in the similarity measure before and after training are shown. We obtain this expectation from $100$ independent rollouts on $100$ independent untrained/trained models, each defined on different random seeds.}
    \label{fig:behavior}
\end{figure}

We address the two central challenges by introducing the mix-and-mask actor-critic: a novel approach to mix and mask latent representations shared between the policy and value function for actor-critic method. We aim to overcome these difficulties by inducing certain desired behaviors within the shared features and modeling a distribution to normalize the two objectives. These behaviors can be seen in Figure \ref{fig:behavior} from the expected difference in similarity $\mathbb{E}[S(h'_\pi,h'_v)-S(h_\pi,h_v)]$ between the policy and value function's hidden representations before and after training. The hidden representations $h'_\pi,h'_v$ were taken from the trained policy and value function, whereas $h_\pi,h_v$ were taken from their untrained counterparts. From this, we see that the separated network approach develops more similar features, especially in mid-level abstractions, whereas the shared backbone network's features diverge, particularly after the shared network. Hence, we present two penalty terms to the objectives to guide the optimization of these mechanisms' parameters to replicate this behavior. We extend these base ideas with auxiliary mechanism representations and contrastive representation learning formulations.

The main contributions of this work is as follows:
\begin{itemize}
    \item [1] We present a novel approach to maintain shared features with actor-critic methods by using mix and mask mechanisms that control the information flow across shared feature spaces.
    \item [2] We introduce new penalty terms to the objective to guide and help these mechanisms induce particular behaviors within the shared feature space.
    \item [3] We consider a novel approach to normalize the objectives using distributional scalarization.
    \item [4] We demonstrate more stable, sample-efficient, and generalized learning with our method on classic control RL tasks.
\end{itemize}


\section{Related Work}
\textbf{Actor-critic methods.} Policy gradient methods define a class of model-free RL optimization algorithms that directly updates the parameters of the policy $\phi_\pi$ with the goal to maximize the expected cumulative reward \cite{reinforce}. However, on it's own, policy gradient optimization is unstable due to their high variance performing gradient estimation \cite{sutton}. To stabilize the optimization, actor-critic methods introduce a value function estimator to assist with the policy's learning by bootstrapping a baseline for policy gradient calculation. In addition, having access to a value function estimator allows greater flexibility to work with different degrees of bootstrapping when using the Bellman Equations. However, these methods do have their limitations. As a function approximation is used to proxy the value function, the policy updates have an asymptotic dependence on the quality of the approximation \cite{sutton}. In general, a wide array of different solutions were proposed \cite{TRPO,PPO,ackter,a3c,acer,SAC,dpg,td3} to improve the actor-critic methods. In this paper, we focus on developing a framework that allows for shared features spaces to obtain more sample-efficient and stable learning.

\textbf{Sharing latent representation.} The idea of sharing latent representation has commonly been practiced in other domains such as vision \cite{vision_tf}, natural language \cite{bert}, and multitask learning \cite{shared_features_mtl}. Often employed using transfer learning, or knowledge transfer across different tasks \cite{transferlearning, transfer_learning_2}, models have often demonstrated and relied on the benefits of this sharing to achieve new level of performance. We aim to extend this idea to deep RL through actor-critic methods.

\textbf{Handling shared actor-critic parameters.} While past works \cite{alphastar,dota2, pinto2017asymmetric} have used shared parameters in some limited fashion, mostly by using a shared backbone, the literature focused on this idea is quite limited. The most relevant work focused on this concept is \cite{cobbe2020phasic}, where they introduce the Phasic Policy Gradient (PPG), which splits the policy and value function's training into two phases: the policy and auxiliary phase. While the policy phase follows standard policy gradient optimization using disjoint policy and value function, the auxiliary phase optimizes an auxiliary loss with a behavioral cloning loss, updating on a set of parameters shared between an auxiliary value function and the policy. In this work, we instead aim to regulate the shared parameters between the policy and value function with the mix and mask mechanisms.

\section{Preliminaries} \label{prelims}
\textbf{MDP Formalization for RL Tasks.}
We consider a MDP $\mathcal{M} = <S, A, P, R, \gamma>$ to describe RL tasks, where $S$ is the set of all states, $A$ is the set of all actions, $P(s_{t+1},r|s_t,a_t)$ is the transition probability function, $r(s_t,a_t)$ is the reward function, and $\gamma$ is the discount factor. Often, as the classic objective of RL is to find the parameters $\phi_\pi$ of policy $\pi_{\phi_\pi}(S|A)$ that maximizes the cumulative reward $G_t = \sum_{k=t+1}^{\infty} \gamma r(s_k,a_k)$, otherwise denoted as the return, we introduce a state-value function $V(s_t)= \mathbb{E}[G_t|s_t]$, or an action-value function $Q(s_t,a_t)= \mathbb{E}[G_t|s_t,a_t]$ under a policy $\pi$ to assess the state's and state-action pair's quality respectively. As to solve these MDPs efficiently, the Bellman equations provides a recursive definition of these value functions.
\begin{gather}
    V(s_t) = \mathbb{E}[r(s_{t+1},a_{t+1}) + \gamma V(s_{t+1}) | s_t]\\
    Q(s_t,a_t) = \mathbb{E}[r(s_{t+1},a_{t+1}) + \gamma \mathbb{E}_{a\sim \pi} Q(s_{t+1},a) | s_t,a_t]
\end{gather}
We use these fundamental ideas of RL in this paper to formulate the mix and mask mechanisms.

\textbf{Cart and Pole (CP) Environment.} The objective of this task is for an agent to balance pole on an un-actuated joint by moving a cart along a friction-less track. The agent is provided the cart's position, cart's velocity, pole's angle, and pole's angle velocity. The action space is defined by a discrete set of forces $\{+1,-1\}$ applied onto the cart, which applies a set magnitude of force. A reward function is defined to provide $+1$ for every timestep that the pole remains upright, and $-1$ once the pole is more than 15 degrees from vertical or the cart moves more than 2.4 units from the center. We consider the task to be solved if the agent is able to achieve a return of $\geq495$ over $100$ independent trials. In this paper, we use this environment to evaluate our methods against prior approaches. We also introduce a new environment, called the non-stationary Cart-and-Pole (nCP), which randomly changes the length of the pole, mass of the pole, and the mass of the cart independently at every trial. The agent will only learn on CP environment, but tested on the nCP environment, as to test generalization.

\textbf{Representation for Actor-critic methods.}
With actor-critic methods, the policy and value function are often represented as neural networks. We can consider various neural architectures depending on the application at hand. In terms of function structure, separate networks are most commonly employed for policy and value function. Some works \cite{alphastar,dota2, pinto2017asymmetric} have used shared backbone with separate heads, often for RL tasks requiring vision. In this paper, we build on these two structures by introducing mechanisms that couple and decouple shared feature spaces.

\textbf{Multi-Objective Optimization.} We construct an RL-specific multi-objective optimization tasks as such. Let a task be defined by a set of functions $F = \{f_1,f_2,\dots\}$, otherwise denoted as the constraint set, and a corresponding set of objectives $O =\{o_1,o_2,\dots\}$.
\begin{equation}
     F^* = \min\limits_F(O(F))
\end{equation}
We consider a solution $F^*$ to be Pareto optimal if $O(f^*_i) \not\geq O(f_i)$,$\forall f_i\in F, f^*_i\in F^*$. Thus, a Pareto solution must take into account the balance over all $o\in O$ as to achieve optimality. A common practice is to use scalarization techniques, which attempt to simplify and map $O$ to a singular objective. This reduction thereby produces an auxiliary single-objective optimization. A classic scalarization method is the weighted sum method, where the singular objective is obtain by some linear combination of $O$. However, for complex tasks, it is usually very difficult to find proper weights to achieve an appropriate balance. Thus, in this work, we reduce the two objectives of the actor-critic by scalarization using distributional weighting, described in Section \ref{algorithim}.


\textbf{A2C Optimization.} The A2C algorithm update the parameters of the policy $\phi_\pi$ and value function $\phi_v$ approximations with the following objective.
\begin{equation}
    \nabla J(\pi, V) = A(s_t,a_t) \nabla_\pi \log(\pi_{\phi_\pi}(a_t|s_{t-1})) - \delta_t \nabla_v  V(s_t) + \mathcal{H}_\pi
\end{equation}
where $A(s_t,a_t)$ is the advantage function and $\delta_t = \mathbb{E}[G_t] - V(s_t)$ is the correction term for the value function. Intuitively, the update seeks to adjust the log likelihood based on the value of the advantage function. To better approximate the advantage function, we use the exponentially-weighted estimator introduced in \cite{gae}, where now $A(s_t,a_t) = \sum_l (\gamma \lambda)^l \delta_{t+l}^v$.  We also add the entropy term $\mathcal{H}_\pi = \log(\pi_{\phi_\pi}(\cdot))$ to the objective to increase policy's exploration \cite{entropy,SAC}. We refer to $\nabla J_\pi = A(s_t,a_t) \nabla_\pi \log(\pi_{\phi_\pi}(a_t|s_{t-1}))$ as the policy update and $\nabla J_v=\delta_t \nabla_v  V(s_t)$ as the value function update. For our experiments, we used a Monte Carlo approach to estimate the return.

\section{Algorithm} \label{algorithim}
In this section, we will discuss our mix-and-mask actor-critic algorithm, which introduces two novel mechanisms: the mix and mask. We found in practice that using them in union requires more temperature parameter tuning, and this variant described later in this section. Thus, we will describe each mechanism first separately, then elaborate on their operation in conjunction. We then describe other alternative variants, such as skip connections, meta-learning and contrastive learning formulations. Lastly, we end this section introducing the distributional scalarization technique and extending the idea of projection of conflicting gradients to this work. The description provided here assumes the actor-critic method requires only a single policy and a single state-value function.

\begin{figure}
    \centering
    \includegraphics[width=\textwidth]{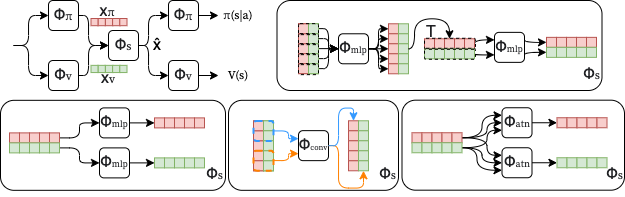}
    \caption{\textbf{Mix Actor Critic}: On the top left, the figure shows the overall architecture of the Mix Actor Critic. On the top right, the figure shows the MLP-Mixer representation of the mix function. On the bottom left, the figure shows the base representation. On the bottom middle, the figure shows the convolution representation. On the bottom right, the figure shows the cross-attention representation.}
    \label{fig:mix_ac}
\end{figure}

\subsection{Mix Actor-critic} \label{mix_actor_critic}
\textbf{Generalized Mix Function.} Let $\phi$ be the parameters for the actor-critic method, and let the representation of the policy and value function be some neural network. First, consider a separate network paradigm, where $\phi_\pi$, $\phi_v$ represent the parameters for the policy and value function, hence making up a partition for $\phi$. We introduce a new set of shared parameters $\phi_s$ for a function $\textit{Mix}(\cdot)$. Suppose $x_\pi$,$x_v$ are latent spaces that can be extracted from the forward propagation of the policy and value function at some intermediate layer. For simplicity, we represent the two latent spaces as matrices of the same rank. Then, we let $\textit{Mix}(\cdot)$ be defined as
\begin{equation}
    \textit{Mix}(x_\pi,x_v) = (\hat{x}_\pi, \hat{x}_v)
\end{equation}
The output of this shared function $\hat{x}_\pi$, $\hat{x}_v$ will then be passed back into policy and value function from where the latent spaces $x_\pi$,$x_Q$ were extracted. Efficiently, we have replaced independent latent spaces from the policy and value functions with new representations that have been coupled. We denote this shared function $\textit{Mix}(\cdot)$ as the mix function. In this paper, we introduce and evaluate on four different representations of the mix function.

\textbf{Mix Function Representations.} The base representation of the mix function uses two independent multi-layer perceptron (MLP) models. The extracted latent representations are concatenated and propagated through each MLP to obtain the mixed representations for the policy and value function. We now extend the mix function to variant representations. The second representation, motivated by the MLPMixer model \cite{tolstikhin2021mlpmixer}, treats each latent space as a channel and pair features from the two latent spaces $x_\pi$,$x_v$ to form a new representation $x = ((x_{(\pi,1)},x_{(v,1)}),(x_{(\pi,2)},x_{(v,2)}),\dots)$. We propagate this representation $x$ through a MLP to obtain a coupled representation. The output is then transposed and passed into another MLP. We can repeat this process, each time increasing the policy and value function's coupling. The third representation stacks the two latent spaces $x_\pi$,$x_v$ together such that the new representation has two distinct channels. We convolve the representation using two independent kernel as the coupling operation. The filters are defined separately for the policy and value function. Again, we repeat this operation to increase the receptive field of the mix function. The final representation we will discuss uses the cross-attention mechanism \cite{DBLP:journals/corr/VaswaniSPUJGKP17}. Given the two latent spaces $x_\pi$,$x_v$, we perform cross-attention in respects to each other. In other words, the mixed policy representation is obtained by passing $x_\pi$ as the key and value, and $x_v$ as the query, while the mixed value representation is obtained by passing $x_v$ as the key and value, and $x_\pi$ as the query. Once each mix function is completed, the mixed representations are placed back to the policy and value functions.

\textbf{Stochastic Mix Function.} We can consider to parameterize the mix function with a distribution $\mathcal{D}$. Let $\mathcal{D}$ be a normal distribution $\mathcal{N}(\mu,\Sigma)$. We decompose the mix function into two mappings $\textit{Mix}_\mu(\cdot), \textit{Mix}_\Sigma(\cdot)$ for the mean $\mu$ and covariance matrix $\Sigma$ respectively. If we represent each feature distribution independently, we let $\Sigma = \sigma^2 I$. For its implementation in practice, we map $\textit{Mix}_\Sigma(\cdot)$ to the log standard deviation. On the other hand, if the feature distribution was multivariate, we must now obtain the full covariance matrix, which is significantly more computationally expensive. Instead of learning a direct mapping, we use the Cholesky factorization by learning a vector $c$ such that $L = (c^Tc) \odot L_I + I$, where $L_I$ is the identity lower-triangular matrix. Thus, we compute the covariance matrix by $\Sigma = L^TL$. We use this approach to softly constrain $\Sigma$ to be positive-definite and non-singular. In future works, we encourage better methods to learn the covariance matrix.

Another consideration we propose is whether $\Sigma$ should be state-independent or state-dependent. For state-independent $\Sigma$, we can store explicit matrices to hold the covariance information. For state-dependent $\Sigma$, we use a MLP to process the state to obtain $\Sigma$.

\textbf{Mix Function Objective.} To guide the mixed representations for the policy $\hat{x}_\pi$, $\hat{x}_v$ towards the desired behavior, we attach a similarity term $\mathcal{S}(\hat{x}_\pi,\hat{x}_v)$ to the objective, such that it only affects the weights of the mixing function. This similarity term provides an incentive to maximize similarity in coupled features.
\begin{equation}
    \nabla_\phi J = \nabla_{\phi_\pi} J_\pi+ \nabla_{\phi_v} J_v + \alpha_s \nabla_{\phi_s} \mathcal{S}(\hat{x}_\pi,\hat{x}_v)
\end{equation}
We introduce the temperature parameter $\alpha_s$ to control the relative importance of having similar mixed representations. We decay $\alpha_s$ through the optimization $\alpha_{(s,t+1)} = \tau*\alpha_{(s,t)}$, where $\tau \in [0,1]$.

\textbf{Similarity Term Representation.} If we assume the mixed representations are sampled from some distribution $\hat{x}_\pi\sim\mathcal{D}_\pi$ and $\hat{x}_v\sim\mathcal{D}_v$, we define our similarity function with a symmetric statistic metric. One choice is the information radius \cite{inforad} $IR(\cdot)$, a symmetric average of the relative Jensen‐Shannon divergence $F(P||Q) = K(P||\frac{P+Q}{2})$, where $K$ denotes the relative information of $P$ and $Q$.
\begin{equation}
    IR(P||Q) = \frac{1}{2}(F(P||Q) + F(Q||P))
\end{equation}
Other similar measures, such as J-divergence $J(\cdot)$ and Arithmetic‐geometric divergence $AG(\cdot)$, the symmetric average of the relative arithmetic‐geometric divergence $G(P||Q) = K(\frac{P+Q}{2}||P)$, were evaluated as well in our experimentation.
\begin{gather}
    J(P||Q) = \frac{1}{2}(K(P||Q) + K(Q||P)) \\
    AG(P||Q) = \frac{1}{2}(G(P||Q) + G(Q||P))
\end{gather}
On the other hand, if we have a deterministic mix function, we use a non-statistical symmetrical distance measure such as cosine similarity.

\begin{figure}
    \centering
    \includegraphics[width=\textwidth]{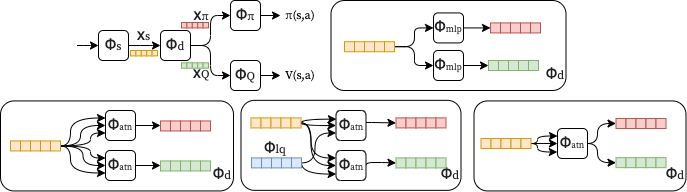}
    \caption{\textbf{Mask Actor Critic}: On the top left, the figure shows the overall architecture of the Mask Actor Critic. On the top right, the figure shows the base representation of the mask function. On the bottom left, the figure shows the self-attention representation. On the bottom middle, the figure shows the latent query self-attention representation. On the bottom right, the figure shows the shared self-attention representation.}
    \label{fig:mask_ac}
\end{figure}

\subsection{Mask Actor-critic}\label{mask_actor_critic}
\textbf{Generalized Mask Function.} We follow the same notation as Section \ref{mix_actor_critic}. However, now consider a shared backbone network paradigm. We define parameters for the policy $\phi_\pi$ and value function $\phi_v$ such that the parameters of the shared backbone function $S$ are defined as $\phi_s = \phi_\pi \cap \phi_v$ and the output $x_s$ of $S$ is propagated into disjoint parameters $\phi'_\pi$, $\phi'_v$. We introduce a new set of parameters $\phi_d$ belonging to a mask function $\textit{Mask}(\cdot)$ such that it serves as an intermediary between the parameters $\phi_s$ and $\phi'_\pi$, $\phi'_v$.
\begin{equation}
    \textit{Mask}(x_s) = (x_\pi,x_v)
\end{equation}
Efficiently, we dedicate a function after the shared function $S$ to decouple the inputs of the policy and value functions before propagating to the disjoint networks. In this paper, we represent the mask function $\textit{Mask}(\cdot)$ in four different manners.

\textbf{Mask Function Representations.} The base representation uses two independent MLP models and propagates the shared latent spaces through the MLPs to obtain the masked representations for the policy and value function. We now extend the mask function to variant representations. The second representation for the mask function consists of two separate self-attention modules, and propagates the shared features to obtain the decoupled representation. The third representation shares the same architecture as the second representation, however, we introduce a learnable latent query that is shared between the self-attention modules as the query variable. The final representation uses a shared self-attention module for both policy and value functions. However, to decouple, we invert the policy attention mask  $a_\pi$ to obtain the value attention mask $a_v$.
\begin{equation}
    a_v = h(1-a_\pi)
\end{equation}
where $h$ is a normalizing constant.

\textbf{Mask Function Objective.} Similar to the mix actor-critic, we introduce a divergence term $\mathcal{D}(x_\pi,x_v)$ to the objective such that its optimization occurs only on the mask function. Unlike the similarity term, the divergence term seeks to maximize dissimilarity in decoupled features.
\begin{equation}
    \nabla_\phi J = \nabla_{\phi_\pi} J_\pi+ \nabla_{\phi_v} J_v + \alpha_D \nabla_{\phi_d} \mathcal{D}(x_\pi,x_v)
\end{equation}
The temperature parameter $\alpha_D$ will control the relative importance having dissimilar masked representations. We follow the same procedure in handling this parameter as the temperature parameter for mix actor-critic.

We apply the ideas of similarity term representation and stochastic mix function from the mix actor-critic and extend them naturally to the stochastic mask function and divergence term representation.

\subsection{Design Choices}
\textbf{Mix-and-Mask Actor-critic.} 
Now we formulate a method for utilizing both the mix and mask mechanisms. We again consider a shared backbone network, and follow the notation from Section \ref{mask_actor_critic}. However, now we introduce the mix mechanism to the disjoint parameters $\phi'_\pi$,$\phi'_v$. Therefore, the objective function now consists both the similarity and divergence term, with each term only affecting their respective mechanism.

\textbf{Shortcut Connections.} From \cite{srivastava2015training}, shortcut connections are a popular approach to ease optimization and avoid performance degradation from additional parameters. We can merge this idea to our mechanisms, passing back the unaltered representations with the altered. In our ablative experiments, we compare dense \cite{densenet} and residual \cite{resnet} skip connections.

\textbf{Auxiliary Mechanisms.} We consider an similar approach to \cite{cobbe2020phasic}, where the outputs of the mix and mask mechanisms are not pass back into the processing network, but rather only used to evaluate the similarity and divergence terms for the objective function. In our case, the optimization of these two penalty terms will affects the preceding parameters of the policy and value function, and the mechanisms themselves are independent of the policy or value function. With this, we can view this formulation as attaching auxiliary mechanisms to the networks.

\textbf{Contrastive Representation Learning.}
We can adapt a contrastive representation learning paradigm to induce the mechanisms' desired behaviors for the mix and mask actor-critic models, tuning these functions instead with a contrastive loss $\mathcal{C}(\cdot)$. To clarify the difference between $\mathcal{C}(\cdot)$ and the penalty terms $(\mathcal{S},\mathcal{D})$, the penalty terms act as soft constraints to the overall policy and value function objective whereas the contrastive loss focuses more on the representation learned and behaves independently from the main objectives. We adopt ideas such as momentum functions $ \textit{Mix'}(x_\pi,x_v),\textit{Mask'}(x_s)$ from \cite{srinivas2020curl}, where their respective parameters  $\phi_\textit{Mix'} = m\phi_\textit{Mix} + (1-m)\phi_\textit{Mix'}$, and $\phi_\textit{Mask'} = m\phi_\textit{Mask} + (1-m)\phi_\textit{Mask'}$, $m\in(0,1)$ are defined as such. For the mix function, we set our anchors $q_\pi,q_k$ to be $\hat{x_\pi}, \hat{x_v}$, the positives $k_+$ to be $\hat{x'_v},\hat{x'_\pi}$ respectively, and the negatives $k_-$ to be $\hat{x'_\pi}, \hat{x'_v}$ respectively to maximize similarity with opposing positives and minimizing individuality with the complementary negatives. For the mask function, we set our anchors to be $\hat{x_\pi}, \hat{x_v}$, the positives to be $\hat{x'_\pi},\hat{x'_v}$ respectively, and the negatives to be $\hat{x'_v}, \hat{x'_\pi}$ respectively to accomplish the opposite of the mask's contrastive objective. We define the contrastive loss as follows:
\begin{equation}
    \mathcal{C}(q,k_+,k_-) = \log \frac{\exp(qWk_+)}{\exp(qWk_+)+\exp(qWk_-)}
\end{equation}
where $W$ is a learnable bilinear product. We follow the logit trick to calculate the cross entropy loss.

\textbf{Distributional Scalarization.}
Consider the cost functions for the policy $J_\pi$ and value function $J_v$ to be represented by some distribution $\mathcal{D}$. In our formulation, we let this distribution be a normally distributed. Then, we obtain the variance from the cost of an untrained policy $\pi_0$ and value function $v_0$. Then, we set these costs $J_\pi(\pi_0)$, $J_v(v_0)$ to have a high z-score, and standardize the objectives using the calculated mean and sampled variance. 

Additionally, we can choose whether or not these distributions are stationary or non-stationary. If they are stationary, we do not update them throughout the optimization process. However, if they are non-stationary, the distributions are updated in intervals. We can aggregate the cost over some interval to readjust the mean and variance set at a lower z-score. However, given the tasks we are evaluating on, we sufficiently demonstrate the efficiency of the distributional scalarization without making the process non-stationary. 

\textbf{Project Conflicting Gradients (PCGrad).}
We adopt ideas from past works \cite{yu2020gradient} to handle gradient conflicts in multi-task learning tasks. In short, the gradients are projected onto the normal plane of the other objective if they are conflicting. The method of projection from one gradient $g_1$ to another gradient $g_2$ is given as such.
\begin{equation}
    g_1 = g_1 - \frac{g_1 \cdot g_2}{||g_2||^2} g_2
\end{equation}
We only enforce this procedure on gradients that update the parameters generating shared feature spaces. In our ablation studies, we compare the distributional scalarization technique with the PCGrad approach.

\section{Results} \label{result}
Here, we provide a comparative analysis between our novel framework and alternative approaches as well as ablative results across different design considerations discussed in Section \ref{algorithim}. For all experiments, we evaluate on the CP and nCP environment with the A2C optimization algorithm described in Section \ref{prelims}and run each network below with a wide range of learning rates and $\epsilon$ values for exponentially-decayed $\epsilon$-greedy action selection.

\subsection{Comparative Evaluation}
\begin{figure}
    \centering
    \includegraphics[width=0.49\textwidth]{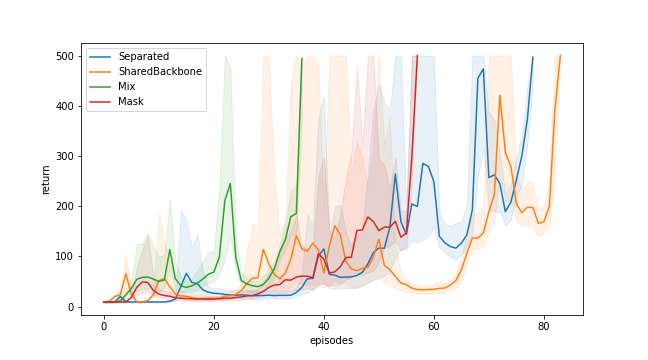}
    \includegraphics[width=0.49\textwidth]{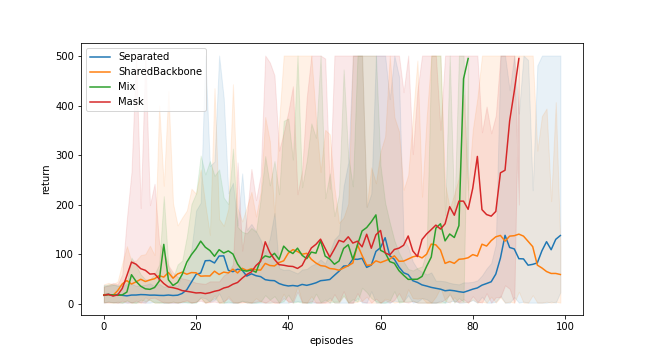}
    \caption{\textbf{CP (left) and nCP (right) Results}: The figure shows the expected return over $100$ independent trials over training. The training is stopped once the task is considered solved or until $100$ episodes were reached.}
    \label{fig:base_compare}
\end{figure}

In this section, we will compare our methods with separated networks and shared backbone networks. In Figure \ref{fig:base_compare}, we show a significant improvement in sample-efficiency and performance, with our methods solving the CP environment. With the methods using shared feature spaces, we do not use distributional scalarization nor PCGrad. The mechanisms learned on fixed temperature values on the penalty terms. The base representations are used for both the mix and mask mechanisms.

However, on the nCP environment, the separated networks and shared backbone networks were not able to fully solve the task within $100$ episodes, whereas our proposed approach were able to. As the figure shows, all of networks exhibit similar trends in the beginning, but we see improvements in only our models. Now, we use the CP and nCP environment to evaluate on the variants and additions we previously introduced in our ablative results.

\subsection{Ablative Results}\label{ablation}
\begin{figure}
    \centering
    \includegraphics[width=0.49\textwidth]{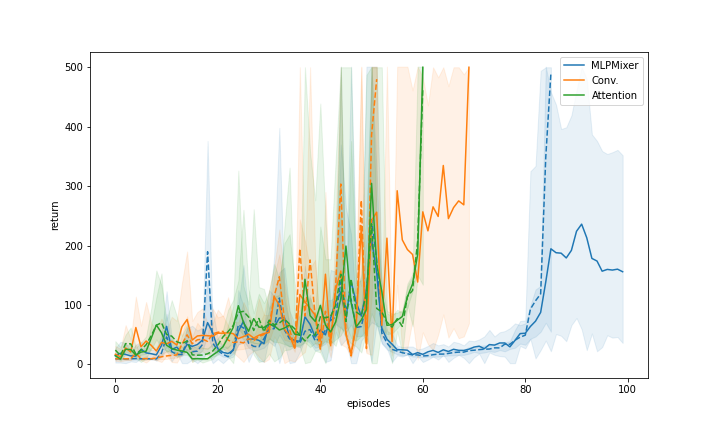}
    \includegraphics[width=0.49\textwidth]{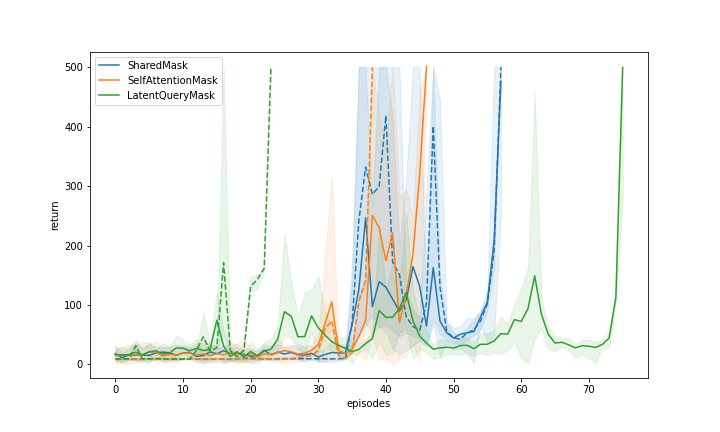}
    \caption{\textbf{Training curve of Mix (left) and Mask (right) representations}: The figure shows the training results on the CP (dotted) and nCP (solid) environments. The training is stopped once the task is considered solved or until $100$ episodes were reached.}
    \label{fig:repr_compare}
\end{figure}
In this section, we evaluate different design choices for the mix and mask actor-critic method and provide empirical results on the CP and nCP environments. First, we test different representations of the mix and mask mechanisms, seen in Figure \ref{fig:repr_compare}. We find that the MLPMixer representation does not converge within $100$ episodes, and exhibits a similar learning curve as the separated network, while the other representations either outperforms or performs comparatively to the base representations. When we compare the training curve between the CP and nCP environments, we conclude that all the networks with the mix and mask mechanisms were able to generalize sufficiently.

On the other hand, we found, through experimentation, that parameterizing the mix or mask functions as a probability distribution fails to converge often with small changes in training hyperparameters, and remains sample-inefficient even when the learning does converge. We notice minimal difference between different probabilistic similarity term representations.

\begin{figure}
    \centering
    \includegraphics[width=0.49\textwidth]{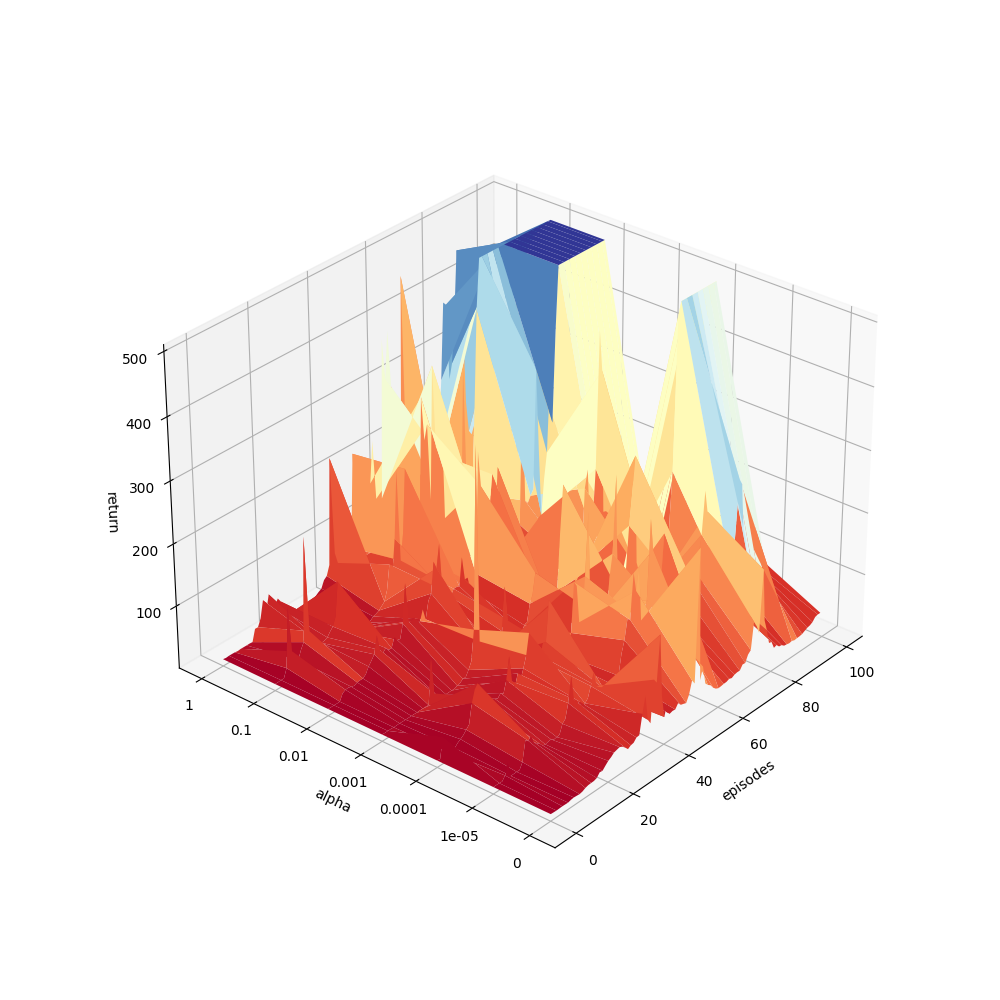}
    \includegraphics[width=0.49\textwidth]{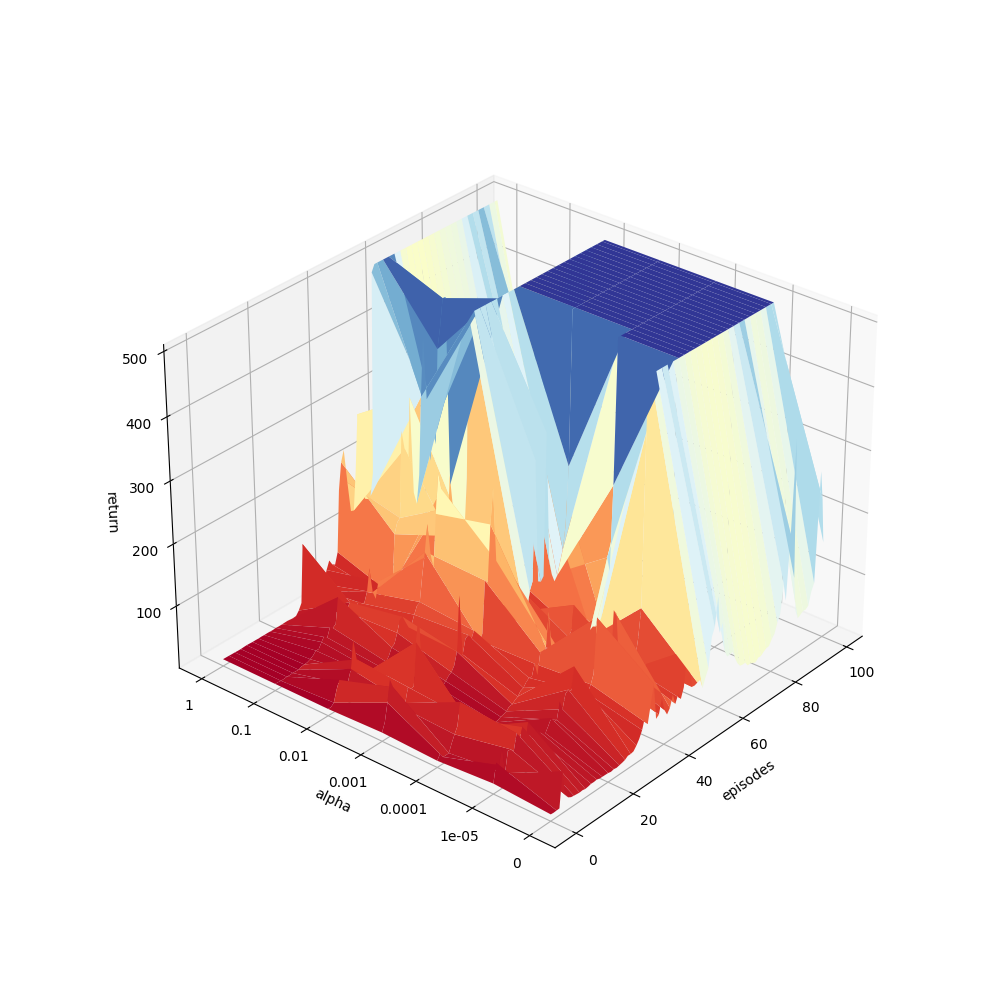}
    \caption{\textbf{Temperature values on penalty terms}: The figure shows the expected return over $100$ independent trials over different temperature values on the penalty terms for mix (left) and mask (right) mechanisms on the CP environment.}
    \label{fig:alpha_compare}
\end{figure}

\begin{figure}
    \centering
    \includegraphics[width=0.49\textwidth]{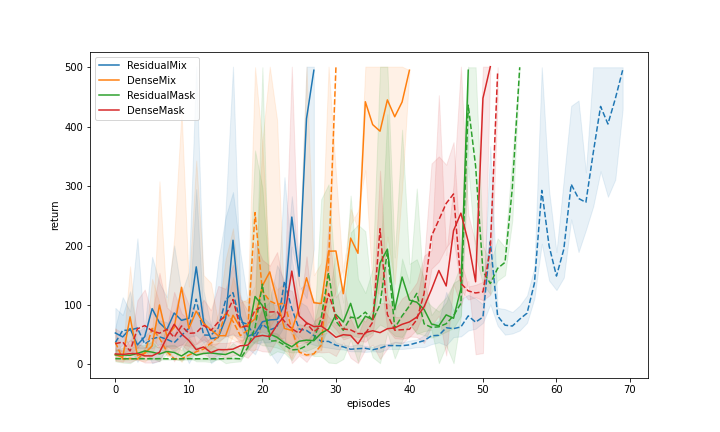}
    \includegraphics[width=0.49\textwidth]{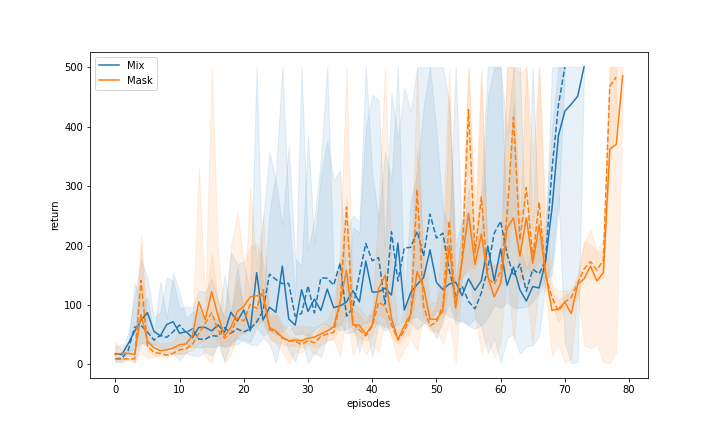}
    \includegraphics[width=0.49\textwidth]{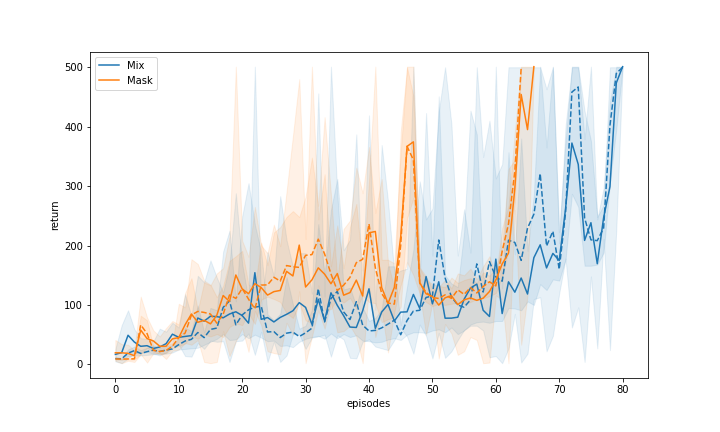}
    \includegraphics[width=0.49\textwidth]{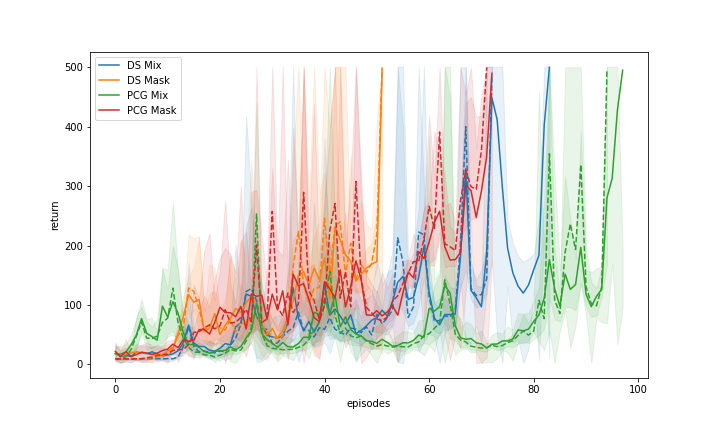}
    \caption{\textbf{Further Ablations on Mix-Mask Variants}: The figures show the following results: skip connections (top,left), auxiliary mechanism (top,right), contrastive learning (bottom,left), and multi-objective handling (bottom,right).}
    \label{fig:variants}
\end{figure}

In Figure \ref{fig:alpha_compare}, we survey different temperature value on the penalty terms imposed on the base mechanisms representations. The experimental results show a need for a sufficiently higher temperature value for the mix mechanisms, which was expected, and a noticeably relaxed requirement for the mask mechanism. This is since, if there is no penalty term imposed on the mask mechanism, the model is equivalent to the shared backbone network. We still see improvements in sample-efficiency and stability once the temperature value increases. However, at a certain point, we begin to see some degradation in performance. This is likely the point at which the penalty term start to overpower the policy and value function objectives. We encourage future work to devise generalized approaches to finding optimal temperature values for the penalty terms.

In Figure \ref{fig:variants}, we explore various design choices and evaluate on the classic control tasks. For each approach, we again use the base representations of the mix and mask mechanisms. We find that skip connections can boost sample-efficiency, as seen obviously by the mix mechanism paired with a dense connection. The performance of the mix mechanism with a residual connection exhibits an interesting behavior as the nCP task is solved before the CP task, which we can only attribute to the experimental noise. The figure also shows that auxiliary mechanisms improves sample-efficiency from the base representations, as does contrastive representation learning. In specific, we see a huge improvement in the mask representation when contrastive learning is employed. Lastly, the figure demonstrates the improvements from using the multi-objective handling procedures: distribution scalarization and PCGrad. From our results, we can see that the distribution scalarization technique allows the networks to converge faster, however the training curve is less stable than PCGrad, which was expected.

\section{Conclusion}
In this paper, we presented the mix and mask actor-critic method and the distributional scalarization technique. We explored promising variants of the base representations of our proposed algorithm, and conducted comparative analysis with ablative studies on various design choices. We encourage future work to extend the use of derived approaches onto different RL tasks and explore different variants of these mechanisms.

\bibliography{reference}{}
\bibliographystyle{plain}

\end{document}